# Hearsay or Fact? How Framing and Elicitation Shape LLM Probability Updates

An-Sheng Jhang
Dongguan University of Technology
ansheng.jhang@gmail.com

**Abstract**

Large language models (LLMs) are increasingly used as judgment assistants and as stand-in participants in psychology experiments, yet little is known about how much their probability estimates move in response to a single new piece of information. We pair premise–hypothesis items from δ-SNLI with one strengthener and one weakener each, elicit the prior and the posteriors in separate calls, and measure the directional movement $s$: the mean posterior change signed by the annotated direction, so that the prior cancels out. We vary whether the update is stated ("Here is something else you learned") or reported ("Someone said that"), and whether the model gives one probability or five weighted estimates. We test seven configurations: two local models (Qwen3-8B, OLMo-2-7B; 240 pairs) and five configurations run through agent command-line harnesses (120 pairs), with hypotheses and analysis rules fixed before the formal runs (not publicly registered). Under stated, direct elicitation all seven move in the annotated direction ($s$ = 18–28 percentage points). Reporting lowers $s$ in all seven point estimates (significantly in six, by 2–5 points) and raises the share of reverse movements in all seven: the effect is consistent but small. Five weighted estimates lower the share of responses at ≤5% or ≥95% in all seven, but in the two local models they also reduce $s$ by 43% and 68%, whereas in the harnessed configurations the point estimates of the change lie between −3.7 and +0.5 points. An elicitation format that reduces extreme values thus need not preserve sensitivity to evidence; the two should be tested separately. Because δ-NLI annotates only direction, we make no claim about how large an update should be. Code, data and the frozen plan: https://github.com/asj252/llm-judgment-studies

## 1 Introduction

Everyday reasoning rarely receives all the information at once. When we see that the ground is wet and passers-by are holding umbrellas, we infer that it has rained; if we then hear that a water-splashing festival is being held nearby, this inference should be weakened. A reasonable reasoner judges not only whether new information makes a conclusion "more likely" or "less likely," but also "by how much"; and the same sentence may also carry different weight depending on whether one knows it for certain oneself or it was reported by someone else.

Large language models (LLMs) are increasingly used as judgment assistants, and even as stand-ins for human cognition that simulate participants in psychology experiments (Vasileiou et al., 2026). Such uses require knowing whether, when a model faces a new piece of information, the probability it reports moves in the right direction, how much it moves, and whether this movement changes with the way the information is presented and the way the question is asked.

Existing research has touched on this problem from several directions, but none has isolated it and measured it.

- **Defeasible-reasoning benchmarks look only at direction.** δ-NLI pairs each premise–hypothesis pair with update sentences that strengthen or weaken the hypothesis (Rudinger et al., 2020); subsequent work treats it as a classification or ranking task, for example requiring only that the updated probability estimate agree with the annotated direction (Wang et al., 2025a). Whether the direction is correct can be evaluated, but the magnitude of the movement has not been reported.

- **Probabilistic NLI and probabilistic inference ask only once.** UNLI collected human subjective probabilities on a 0–1 scale for SNLI items (Chen et al., 2020); ProbCOPA compares the distributions of humans and reasoning models on a 0–100 scale and finds that the models rarely give mid-range values and lean toward the two ends (Kamath et al., 2026). These works all involve a single judgment, with no comparison of "before and after new information appears."
- **Research on forecast updating does not control direction.** EvolveCast has models report a forecast probability once before and once after seeing a news item, with Metaculus community forecasts as the reference (Yuan et al., 2025). The news is obtained by retrieval, its direction is not controlled, and a single news item may also mix several events.

In this paper, we measure the probability movement of LLMs under **single-sentence updates whose direction is fixed in advance**, and test the effects of two manipulations.

**First, reported wording.** People adjust how much they accept information according to the credibility of its source (Hovland & Weiss, 1951; Pornpitakpan, 2004). For LLMs, research on sources has mostly concerned knowledge conflicts and answer selection: models are given two mutually conflicting documents labeled with different sources to see which one they choose (Schuster et al., 2026), or are tested on whether they defer more to an answer labeled as coming from an "expert" (Bajaj & Tiganj, 2026). In linguistics, the extent to which "X said that p" licenses the inference of p is a classic question in research on verb veridicality (Ross & Pavlick, 2019; Jiang & de Marneffe, 2019), but there the dependent variable is an entailment label or speaker commitment, not the probability movement of another hypothesis. As far as our search found, no study has presented **the same new piece of information** in two wordings, "Here is something else you learned: U" and "Someone said that U", and compared the models' probability movement for the same hypothesis.

**Second, extreme values and multi-estimate elicitation.** When a model is asked directly for a single probability, its answers tend to crowd at the two ends: verbalized confidence mostly falls in 80–100% (Xiong et al., 2024), and reasoning models rarely give mid-range values (Kamath et al., 2026). For research on updating, this is not merely a calibration problem: if the prior is already reported as 100%, strengthening information has no room to move it upward, and whether the model was influenced by the sentence cannot be measured. Verbalized Sampling (VS) proposes that, instead of having the model give only one answer, the model list several candidates at once, each labeled with a probability, to alleviate the problem of models always giving the most typical answer (Zhang et al., 2025); research on question-answering calibration has also found that giving several candidates at once, or a verbalized probability distribution, can reduce calibration error for some models and datasets (Tian et al., 2023; Xiong et al., 2024; Wang et al., 2025b). As far as our search found, we have not yet found a study that, in single-sentence, direction-controlled probability updating, tests the effects of such elicitation formats on both endpoint values and directional sensitivity.

We sample 240 premise–hypothesis pairs from δ-SNLI (240 distinct images), take one strengthener and one weakener for each pair, and collect probabilities from seven model configurations: Qwen3-8B (thinking off) and OLMo-2-7B-Instruct run locally (240 pairs each); the commercial model gpt-6.1-sol called through the codex command-line interface (120 pairs); and, in an extension run with the same protocol, gpt-oss-120b, claude-sonnet-5.5 and gemini-3.8-flash at two reasoning efforts, called through Google's agy command-line interface (the same 120 pairs). For each item, the model reports, in separate calls, the prior and the posteriors under two presentations: stated ("Here is something else you learned") and reported ("Someone said that"). There are two elicitation formats: reporting a single probability directly (direct elicitation), or listing five estimates, each labeled with a weight (multi-estimate elicitation). The main measure is the **directional movement** $s$: the average of the increase under strengtheners and the decrease under weakeners, in which the prior cancels out. The hypotheses, the smallest effect sizes of interest (SESOIs; 5 points for probability movement and 10 percentage points for differences in endpoint rate), the handling of missing data, and the statistical methods were frozen before the formal set was run and kept unchanged for the extension; the statistics use a cluster bootstrap (resampling images) and Holm correction within each configuration.

Our main results are:

1. **All seven configurations move, but not every time.** Under stated presentation and direct elicitation, the directional movement is between 17.6 and 28.3 points; but counted per update sentence, 16%–29% of the update sentences did not change the reported probability for the local models, and 7%–15% for the configurations accessed through agent harnesses.
2. **Reporting shifts the direction of movement, a little.** When the update is changed to "Someone said," the point estimate of the directional movement becomes smaller in all seven configurations (by 0.9 to 5.4 points; the direction is confirmed in six), while the proportion of movements in the reverse direction rises in all seven (for example, the reversal rate of OLMo-2 goes from 0.16 to 0.22). This cannot simply be read as "discounting" in the same direction. Only the point estimate of codex reaches 5 points.
3. **Multi-estimate elicitation de-extremizes; in the local models it also desensitizes.** The proportion of reported values at the endpoints (≤5% or ≥95%), computed with the pre-specified condition weights and with the weighted mean as the reported value under multi-estimate elicitation, drops in all seven configurations (e.g., from 30%, 42%, and 25% to 4%, 5%, and 14% in the original three); the directional movement of the two local models also decreases by about 43% and 68%, while in the five harnessed configurations the point estimates lie between −3.7 and +0.5 points.

Our contributions are:

1. **A controlled update paradigm and materials**: single-sentence updates with annotated direction, with strengtheners and weakeners in pairs; two numerical probabilities for the same hypothesis, before and after; a directional movement measure in which the prior cancels out; and worst-case imputation bounds for missing units. The main design and analysis rules were frozen before the formal run; the only deviation after the formal run of the original three configurations (Addendum A1), and the events during the extension, are described in the Methods section.
2. **A test of reporting wording**: when the same information is changed to an anonymous report, the models' movement becomes smaller and reverse movement increases; this pattern is consistent in direction across seven configurations, small in size, and differs across configurations when broken down by update type.
3. **Multi-estimate elicitation applied to updating**: we test the format "list five estimates and weight them" in single-sentence, direction-controlled probability updating, find that while reducing endpoint values it can substantially reduce directional sensitivity to evidence in some models but not in others, and raise the issue that "de-extremizing" and "desensitizing" need to be tested separately.
4. **A behavioral diagnostic for elicitation formats**: because the protocol fixes the direction of the evidence and measures prior and posterior separately, it yields a per-model response profile to an elicitation format: how much the format reduces extreme values, what it costs in sensitivity to evidence, and whether the spread of the elicited candidates behaves like an expression of uncertainty. Across seven configurations these profiles differ markedly, which suggests that formats such as Verbalized Sampling should be checked model by model before their outputs are read as calibrated judgments.

We do **not** claim that the models "under-update" or "over-update": δ-NLI annotates only direction and provides no human reference for the appropriate magnitude. A human probability reference under the same paradigm is left for future work.

—

# 2 Related Work

## 2.1 Defeasible reasoning and probabilistic natural language inference

δ-NLI casts defeasible reasoning as a task over (P, H, U) triples: given a premise P and a hypothesis H, judge whether an update U makes H more likely (strengthening) or less likely (weakening). The data comprise three subsets, SNLI, ATOMIC, and Social Chemistry, validated by human annotators on a 5-point Likert scale; no numerical probabilities were collected (Rudinger et al., 2020). On δ-SNLI, GPT-4o already reaches a direction-classification accuracy of 92.6%, higher than the human accuracy reported in the original paper (Srikanth & Rudinger, 2025).

UNLI relabels a portion of SNLI with 0–1 subjective probabilities: each pair is rated by 2–3 people with a slider and the ratings are averaged, with an inter-rater correlation of about 0.70 (Chen et al., 2020). Always Tell Me The Odds trains probability estimation on UNLI and then uses δ-NLI to constrain the direction of the estimates, evaluating with ranking accuracy on δ-SNLI; it does not report the probability difference before and after the update for the same item (Wang et al., 2025a). In a pilot study, BIRD found that it was difficult to reach consensus when people were asked to report probabilities directly, and therefore switched to pairwise preference judgments (Feng et al., 2025). ProbCOPA collects 0–100 ratings from 25–30 people for each of 210 COPA items and finds that human distributions are graded, whereas 8 reasoning models lean toward the two ends; there is no significant difference between high and low thinking budgets; and adjusting temperature or persona prompts has limited effect (Kamath et al., 2026).

Another line of work studies belief revision when a rule is overturned by a counterexample. Belief-R uses three-option multiple-choice items to examine whether models can retract a conclusion after a new premise appears (Wilie et al., 2024). Vasileiou et al. compare humans and three frontier models on everyday scenarios of the form “rule + instance + counterexample”; the dependent variables are the proportion of revising the rule versus doubting the instance, and whether the revision generalizes to related beliefs; the proportion with which the models target the refuted rule is similar to that of humans, but they almost never generalize (Vasileiou et al., 2026). These works are concerned with “which one to revise,” not with the magnitude of probability movement.

**Difference from this paper**: we adopt the direction annotations of δ-NLI, but we measure the numerical probability of the same hypothesis before and after the update and its movement, rather than direction classification or a single probability.

## 2.2 Probability updating in LLMs

EvolveCast is closest to our work: on Metaculus questions, it has models report a probability once before and once after a news item, obtaining 1,613 question–news pairs, and uses the community forecast’s values at the two time points as the reference (Yuan et al., 2025). The differences are that the news is obtained by retrieval and its direction is not controlled; there is no manipulation of source or wording; and elicitation formats are not compared. The authors state that the models’ update magnitudes tend to be too small, but the main text reports metrics such as MSE and direction agreement rate and does not directly compare the magnitude of movement between models and humans.

Other work examines updating from the perspective of Bayesian coherence. Imran et al. read priors, likelihoods, and posteriors from token log-probabilities (rather than verbalized probabilities), use a Bayesian Coherence Coefficient (BCC) to measure pretrained models from five families without instruction tuning, and find that the coefficient is highly correlated with the logarithm of the parameter count ($r = 0.906$) (Imran et al., 2025). Kim et al. compare models’ verbalized confidence on SciQ under no-evidence and with-evidence conditions; one of their manipulations of evidence strength is source credibility (a peer-reviewed paper published by a university professor vs. an anonymous blog post); all of their evidence supports the correct answer, and there is no weakening direction (Kim et al., 2025). LEAP has models give a likelihood for each piece of evidence separately, and then combines them into a posterior using an explicit prior and a deterministic model; the posterior is not self-reported by the model (Chen et al., 2026).

**Difference from this paper**: in this paper, the new information is a single sentence with annotated direction, strengtheners and weakeners come in pairs, and the same model self-reports two probabilities, before and after, for the same hypothesis.

## 2.3 Sources, reporting, and evidential markers

**Linguistics: verb veridicality and speaker commitment.** "X said that p" and "X knows that p" commit to p to different degrees, a classic difficulty for NLI. Ross & Pavlick test NLI models with 1,500 sentence pairs and 137 verbs and find that both humans and BERT tend to treat verbs as veridical (Ross & Pavlick, 2019); CommitmentBank annotates how speaker commitment varies in negation, modal, question, and conditional environments (de Marneffe et al., 2019), and Jiang & de Marneffe evaluate NLI models on it (Jiang & de Marneffe, 2019); embedding non-factive verbs or evidential adverbs also disrupts the entailment judgments of LLMs (Basmov et al., 2024). The dependent variable in these works is an entailment label or a commitment rating, and they ask "whether p can be inferred from the report"; we ask "how a reported U moves the probability of another hypothesis H."

**Epistemic and evidential markers in prompts.** Adding markers such as "I've heard it's" or "I'm sure" to a question changes the answer accuracy of models (Zhou et al., 2023). Belém et al. use the template "Speaker believes it is [uncertainty expression] that S" to have models report a 0–100 probability from the speaker's perspective, comparing 14 uncertainty expressions; the study uses only wordings attributed to the speaker and has no non-attributed control (Belém et al., 2024). Du et al. classify belief expressions into 17 types (including evidentiality) and examine whether models follow the context or their own priors (Du et al., 2026).

**Source credibility and LLMs.** In knowledge conflicts, models prefer institutional sources over individuals and social media, but this preference reverses once a claim from a low-credibility source is repeated multiple times (Schuster et al., 2026). That study presents two conflicting documents labeled with sources at the same time and measures preference by option-token probabilities, without having the model self-report probabilities. Models defer more readily to a previous answer labeled as coming from a "human expert" (Bajaj & Tiganj, 2026); in a large-scale source-discernment test, the models' ability to discern source reliability is close to chance, and their reliance on source popularity is about twice their reliance on reliability (Ashkinaze et al., 2026); in a Turkish evidential-suffix task, humans show a stable source-credibility effect, whereas the effect in LLMs is unstable and often reversed (Karakaş & Şimşek, 2026).

**Human research.** People discount untrustworthy sources at the time they receive the information, and this discount fades over time (Hovland & Weiss, 1951); for a review of fifty years of source-credibility research, see Pornpitakpan (2004). Harris et al. model the expertise and trustworthiness of experts with Bayesian networks and test the model with numerical ratings of proposition credibility (Harris et al., 2016).

**Difference from this paper**: existing LLM studies manipulate source identity, source preference, or uncertainty expressions, and their dependent variables are mostly answer selection, accuracy, or a single probability. We hold the content of the new information fixed and change only "learned" to "someone said," comparing the probability movement of the same hypothesis in both the strengthening and the weakening directions; our results show that this change not only makes the movement smaller but also increases reverse movement (Section 4.3).

## 2.4 Verbalized probability, extreme responses, and multi-estimate elicitation

The verbalized confidence of LLMs is often too high: Xiong et al. find that confidence mostly falls in 80–100% and is mostly a multiple of 5; having the model give Top-K guesses with their probabilities reduces calibration error (Xiong et al., 2024). Tian et al. find on GPT-4 that giving 4 candidates with probabilities at once yields a lower ECE than giving only 1 on SciQ and TruthfulQA, but a higher one on TriviaQA (Tian et al., 2023). Having the model verbalize a probability distribution over candidate answers that sums to 1 also lowered the Brier score and ECE on several closed-ended datasets, with smaller models benefiting more, but there are also tasks where the effect was worse (for example, on the mathematical reasoning dataset OlympiadBench, both the Brier score and the ECE of Qwen3-4B with the distribution method are worse than with verbalized confidence) (Wang et al., 2025b). Other research treats listing multiple candidates at once, each with a confidence, as a baseline and finds that the candidates' confidences often sum to more than 1 and that overconfidence remains (Wang & Stengel-Eskin, 2026). In opinion-distribution tasks, distributions reported directly in words are closer to population distributions than distributions read from log-probs (Meister et al., 2025).

Verbalized Sampling has a model generate k responses at once, each labeled with a probability, to alleviate the "mode collapse" in which aligned models always give the most typical answer, and it mainly evaluates generation diversity (Zhang et al., 2025). Its appendix examines the calibration of verbalized probabilities: on closed answer spaces such as state names, countries, and programming languages, it compares the correlation between each answer's verbalized probability and its corpus frequency. The calibration target there is "the relative frequency with which each answer occurs," not "the probability that an event holds"; endpoint values are not counted, and there are no two judgments, before and after.

Psychology has related approaches: having people assign probabilities to every interval across the whole range of values reduces overprecision (Haran et al., 2010); averaging two estimates from the same person is more accurate than a single estimate (Vul & Pashler, 2008); assuming that the first estimate was wrong before giving a second estimate yields lower error after averaging (Herzog & Hertwig, 2009); in two-alternative judgments, averaging the same person's two confidence ratings improves the accuracy of confidence judgments more than taking the higher of the two (Litvinova et al., 2020). However, the two estimates in these "inner crowd" studies are separated in time and relatively independent, whereas multi-estimate elicitation produces them in a single generation with candidates that depend on one another, so the two should not be directly equated.

**Difference from this paper**: existing work applies multi-candidate or verbalized-distribution elicitation to the calibration of one-shot question answering or to generation diversity; we apply it to before-and-after judgments of "the probability that something holds in a scenario", and test its effects on both endpoint values and directional sensitivity. We do not require the five estimates to differ from one another, so as not to create dispersion artificially; if a model gives five nearly identical estimates, that is itself a result.

—

# 3 Methods

## 3.1 Materials

**Source.** Items were taken from the SNLI (Bowman et al., 2015) subset of δ-NLI (δ-SNLI; Rudinger et al. (2020) github.com/rudinger/defeasible-nli, commit c675ffc), merging dev and test, for a total of 3,860 rows. The data inherit SNLI's CC BY-SA 4.0 license. Each row consists of a premise P (an image caption), a hypothesis H, and a human-written update U, labelled as a strengthener (S)

or a weakener (W). We do not use δ-ATOMIC (whose hypotheses are mostly template fragments) or δ-SOCIAL (which has no premise sentence).

**Item selection.** We first excluded the 238 rows marked as "impossible to write an update", leaving 3,622 rows. Aggregating by premise–hypothesis pair, we kept only pairs that have both an S and a W update: 194 pairs in dev and 202 in test. We further excluded pairs that share images with the two human-check materials (32 pairs), yielding a candidate pool of 364 pairs from 357 images. We shuffled the pool with seed 20261070 and took pairs in order, taking each image only once; for each pair, the same random number generator was then used to select one S and one W update (for replacement pairs and for the sentence selection in Block B, an independent generator seeded with the pair's ID was used instead, so that other pairs are not disturbed). The repeat subset was drawn from Block A with seed 20261071. In total we took 270 pairs: 30 for the development set and 240 for the formal set, the two being disjoint by image. The formal set is divided into two blocks of 120 pairs each; Block A is the originally planned formal set, and Block B was drawn further down the same shuffled sequence when the sample was expanded in v0.7 according to power scenarios.

**Exclusion of items involving protected attributes.** If the hypothesis H or a selected update S or W asserted or asked about a person's race, ethnicity, nationality, or religion, the whole pair was excluded and replaced by the next eligible pair in the original sampling sequence; pairs not excluded, and their selected sentences, were left unchanged (the premise P was not within the scope of this check). Candidates were first screened with a frozen keyword list; each flagged pair was then judged individually, with the reason recorded, by the language-model agent responsible for material checks in the research workflow, and the final list was re-checked with a broader word list. This was a material check performed by a model, not human screening. In total, 8 pairs were excluded. Rationale: in the development-set smoke test, codex refused to answer all such items; moreover, asking a model to infer race from appearance is itself ethically problematic.

**Clusters.** We use SNLI images as clusters: different captions of the same image are treated as the same cluster. The 240 pairs of the formal set come from 240 distinct images.

**Direction spot check.** We additionally drew 20 premise–hypothesis–update items (10 strengtheners and 10 weakeners, included in neither the development set nor the formal set), presented only in the stated version. The author, as the sole human rater and without seeing the labels, judged whether each update was a strengthener, a weakener, had no effect, was incomprehensible, or was ungrammatical: 17 items agreed with the label, 2 were judged "no effect", and 1 was in the opposite direction. This serves only as a check on the materials and is not used as human agreement-rate data (the original δ-NLI paper reports a human accuracy of 83.6% on the SNLI subset).

## 3.2 Item format and elicitation

All prompts are in English and consist of a single user message with no system prompt. The item format is:

```
Here is a situation: P
[Here is something else you learned: U]  (stated)
[Someone said that u]                    (reported)

Using common sense, how likely is it that the following statement is true?
Statement: H

(elicitation-specific ending)

Do not use any tools; answer directly.
```

Prior items omit the second line. In the reported version, u is U with its first letter lowercased according to a frozen rule: the first letter is left unchanged if the sentence begins with a non-letter, is already lowercase, begins with "I" or "I'", begins with an abbreviation containing capital letters, or begins with a word in a fixed proper-noun list (Batman, Christmas, Sally, Gone, Mom; determined from the sentence-initial words of all 3,622 updates, independently of which items were sampled);

no general named-entity recognition is performed. If P, H, or U lacks sentence-final punctuation, a period is added; all other text (including spelling errors) is left unchanged. "Do not use any tools" is required for codex isolation; all configurations use the same prompt.

The endings for the two elicitations:

- **Direct elicitation**: "What is the probability that the statement is true?", with the final line required to be written as `Probability: NN%` (0–100, decimals allowed).
- **Multi-estimate elicitation** (inspired by Verbalized Sampling; Zhang et al. (2025)): "Give 5 probability estimates that the statement is true, each with a weight for how likely you think that estimate is; the weights sum to 1.", with the answer required to end with 5 lines of the form `Estimate k: NN% | Weight: W`. The aggregated value is the weighted mean.

Each pair has 10 conditions under each configuration: prior $\times$ 2 elicitations, plus posterior $\times$ 2 update types $\times$ 2 presentations $\times$ 2 elicitations. Within the same elicitation, the stated and reported versions share the same prior. A further 20 pairs were drawn as the repeat subset, for which every condition was asked once more independently, in order to estimate between-call variability.

## 3.3 Models tested

| Configuration | Model | How run | Key settings |
|---|---|---|---|
| Qwen3 (thinking off) | Qwen3-8B (HF commit b968826) | Local llama.cpp b11560, GGUF Q8_0 | Chat template `enable_thinking=False` context 16,384, at most 8,192 generated tokens |
| OLMo-2 | OLMo-2-1124-7B-Instruct (HF commit 470b1fb) | Same as above | Context 4,096; generation up to the end of the context |
| codex | gpt-6.1-sol, reasoning effort medium | codex CLI, ChatGPT subscription quota | Invoked through the codex CLI harness; isolation and inclusion rules below |

The two local configurations use the same sampling parameters: temperature 0.6, top-k 20, top-p 0.95, min-p 0, with all other samplers disabled; prompts are rendered with the chat template bundled with the HF tokenizer, as a single user message. Every call uses an independent context. The call seed is the first 8 hexadecimal digits of the SHA-256 hash of the string "{set}|{configuration}|{unit id}|a{attempt}|{base seed}" (base seed: 20261080 for the development set, 20261081 for the formal set); the run order is shuffled with "base seed $\times$ 10 + configuration offset (qwen3-off 0, olmo2 2, codex 3)". A generation that used up the length limit before the end-of-sequence token was classified as truncated (this did not occur in the formal run).

codex is a commercial model invoked through an agent harness (codex CLI 0.162.0-alpha.17.2; in the formal run it was executed with 3 parallel worker processes, with the same attempt and resampling rules for each unit as in sequential execution). We cannot control its system prompt, and we did not set its sampling parameters or seed (the call seed described above has no effect for codex); our conclusions apply only to "the model inside the harness". Each call was made in an isolated temporary directory, and the event stream and internal logs were checked and handled according to the following rules:

- **Tool contamination**: if the logs or event stream contain a record of tool use (confirmed), or show multiple samples, multiple answers, or input tokens exceeding 1.5 times the baseline (7,300 plus one third of the prompt length in characters) (suspected), the attempt is voided and the same condition is resampled once; if it is still contaminated after resampling, it is recorded as `tool_contaminated` (missing).
- **Disconnection**: when the only integrity problem is that the message counts in the logs and in the events disagree, and the difference can be explained by the number of disconnection retries in the logs, only that attempt is voided and resampled, without counting against the resampling allowance above; the run stops only after three consecutive disconnections on the same unit.
- **Call errors** (non-zero exit, timeout, unknown event, no final answer, etc.) and **quota-exhaustion signals**: the run stops; the session is not voided, and the work is redone when the launcher script resumes the run. Addendum A1 (after the formal run had started) also placed in this category calls that "could not connect to the server and left no sampling record or output".
- **Session invalidation**: a change in the global instruction file or any other log-integrity error voids the entire session; voided sessions and sessions that did not terminate normally are not included in the analysis.

Addendum A1 changed codex's experiment fingerprint (a hash binding the plan, sampling, prompts, parser, runner and materials), so both the development set and the formal set for codex were rerun under the new fingerprint, and the earlier results do not enter the analysis. The isolation parameters and the frozen harness settings are recorded in the run metadata.

**Extension configurations (v2).** Four further configurations were added after the original study: gpt-oss-120b (reasoning effort medium), claude-sonnet-5.5 (medium), and gemini-3.8-flash at two reasoning efforts (low and high). All four were run through Google's Antigravity command-line agent harness (`agy`, version 1.4.0, with automatic updates disabled), using the subscription quota; the model and reasoning effort are selected by the model name passed to the harness. As with codex, we cannot control the system prompt, sampling parameters or seed, and conclusions apply only to "the model inside the harness".

| Configuration | Model | How run | Key settings |
|---|---|---|---|
| gpt-oss-120b | gpt-oss-120b, effort medium | agy CLI 1.4.0 | Custom agent declaring `tools: []` sandbox; one call per fresh conversation |
| claude-sonnet-5.5 | claude-sonnet-5.5, effort medium | Same as above | Same as above |
| gemini-3.8-flash (low) | gemini-3.8-flash, effort low | Same as above | Same as above |
| gemini-3.8-flash (high) | gemini-3.8-flash, effort high | Same as above | Same as above |

Each call started the harness in a new, empty temporary directory outside any repository, with a single custom agent definition that declares `tools: []`, excludes the harness's default components, and contains the single instruction "Answer the user's question directly."; in probes, the models still reported a few residual tools (documentation retrieval from an installed plugin and resource-listing tools), so the declaration is not by itself proof that no tool is available, and isolation rests on the declaration together with the event-stream check below; the prompt (identical to the one used for the other configurations, including the final "Do not use any tools; answer directly.") was passed on standard input and the full event stream was saved. The inclusion rules mirror those for codex: any step in the event stream other than the user input and the final answer (including any tool event, denied action, or unknown step type) marks the attempt as contaminated, and it is discarded and resampled; timeouts and disconnections are discarded and resampled; three consecutive attempts of the same kind (contaminated, disconnected or truncated) on the same unit stop the run for review. Parse failures are resampled once, as for the other configurations. One normalization was added before parsing, fixed after the neutral probes and before any study item was run: Unicode space separators (category Zs, e.g. non-breaking spaces) are replaced by ordinary spaces. In the formal set this changed the input for 501 of 1,400 gpt-oss-120b records, 7 of which parsed only after normalization; it changed no record of the other three configurations. Before and after every call, the harness version, a hash of the permission settings, the set of settings keys, the hash of the agent definition and the model name reported by the service were compared with a baseline taken at the start; any change stopped the run before the next unit. Rate limits were handled by exponential back-off without discarding completed units.

The extension used the 30 development pairs and the 120 Block-A pairs of the formal set (the same pairs as codex), with the same 20-pair repeat subset: 300 development units and 1,400 formal units per configuration. Each configuration had to pass the same development-set acceptance check as the original configurations before its formal run.

## 3.4 Parsing and missing data

Following the prompt's instructions, parsing reads the "last" answer (parsing rule v2). For direct elicitation, the last `Probability:` field is taken; if that field has no valid number on the same line, parsing fails, without falling back to earlier fields. For multi-estimate elicitation, the final group is taken starting from the last line that looks like "Estimate 1" (lenient match), but all estimate lines within the group must still strictly qualify: exactly 5 lines, numbered 1–5, each value in 0–100 with a percent sign, and weights in [0, 1]. As long as the weights sum to a positive number, they are normalized by that sum (the weighted mean is invariant to overall scaling); whether the sum falls within [0.95, 1.05] is reported only descriptively.

Parsing rule v2 was revised after inspecting the types of format failure in the development-set smoke test (for example, an extra `Probability:` field written in a heading, estimates listed once in the body and again at the end, or weights not summing to 1). No probability values were looked at during the revision. The Results section also reports the parse rate under the earlier parsing rule v1.

When parsing fails or the output is truncated, the same prompt is resampled once (this allowance is counted separately from the resampling allowance for tool contamination); if it still fails, the unit is recorded as missing. For each unit, the first qualifying attempt is used. Each comparison uses only the items for which all conditions involved in the comparison are valid (the common complete set). We also compute a missing-data bound: the contributions of missing units are replaced by the value least favourable to the conclusion (0 or 100 for linear quantities, 0 or 1 for endpoint indicators), and the bootstrap is rerun; only if this conservative interval still supports the conclusion do we use the verdict "the adjusted bound exceeds the SESOI (worst-case imputation of missing units)".

## 3.5 Measures and analysis

- **Directional movement**: $\Delta = p_1 - p_0$ (posterior minus prior), and the directional movement $s = \Delta \times$ direction (+1 for S, −1 for W), averaged over the S and W updates of each pair; $s > 0$ indicates movement in the annotated direction. The prior cancels out in s, so s depends only on the two posteriors. We also report Δ separately for strengtheners and weakeners, the difference between the two posteriors $D = p_S - p_W$, the agreement rate, the reversal rate, and the zero-movement rate.
- **Endpoint rate**: the proportion of reported values ≤ 5% or ≥ 95%; for H4a it is weighted 1/2 for the prior, 1/4 for the stated strengthener, and 1/4 for the stated weakener.
- **Confirmatory hypotheses** (Table 1): H1, directional sensitivity; H2, reported wording; H4a, endpoint rate under multi-estimate elicitation; H4b, directional sensitivity under multi-estimate elicitation.

Table 1: Confirmatory hypotheses

| Hypothesis | Statistic | Sidedness | Smallest effect size of interest (SESOI) |
|---|---|---|---|
| H1 | Mean s, stated, direct elicitation | One-sided, > 0 | 5 points |
| H2 | Mean s, stated minus reported (direct elicitation) | One-sided, > 0 | 5 points |
| H4a | Weighted endpoint rate, direct minus multi-estimate | One-sided, > 0 | 10 percentage points |
| H4b | Mean s, multi-estimate minus direct (stated) | Two-sided | 5 points |

- **Statistics**: items are weighted equally. We use a cluster bootstrap (each replicate resamples clusters, carrying along all conditions of each cluster), with B = 5,000, seed 20261090, and non-interpolated quantiles; we report two-sided 95% intervals. Bootstrap p-values: the one-sided p-value is the proportion of resampled statistics falling on the side opposite to the predicted direction (including zero); the two-sided p-value is twice the smaller of the two one-sided proportions (capped at 1). For each configuration, {H1, H2, H4a, H4b} form one Holm family: hypotheses are ordered by p-value and compared step by step against $\alpha/(4-k)$, stopping at the first non-rejection; the adjusted bounds are per-hypothesis adjusted intervals, not a joint confidence region. Each configuration is tested on its own planned sample (240 pairs for the local configurations, 120 pairs for codex); cross-configuration comparisons are descriptive only, and we additionally report the results of the local configurations on the same 120 pairs used for codex.

- **Verdict wording** (frozen in advance): "the adjusted bound exceeds the SESOI" = the Holm-adjusted bound nearer to zero strictly exceeds the SESOI; "the point estimate reaches the SESOI, but the adjusted bound does not" = the point estimate is ≥ the SESOI, but that bound does not exceed it; "direction confirmed; point estimate below the SESOI" = the direction is confirmed (that bound lies beyond zero), but the point estimate is smaller than the SESOI. For H1, H2, and H4a we use the adjusted one-sided bound; for H4b we use the end of the adjusted two-sided interval nearer to zero. H2 has one additional pre-specified interpretation rule: if the reversal rate under reported presentation is higher than under stated presentation, the conclusion is worded as "reporting shifts the direction of movement".
- **Pre-specified sensitivity analysis**: excluding pairs whose update is quoted speech (an update that reports the content of someone's utterance, so that its reported version becomes a quoted-speech update, i.e., speech nested within a report), we recompute H1, H2, and H4b, keep H4a on the original sample, and still use the four-hypothesis Holm family. The frozen list of quoted-speech updates has 11 IDs, of which 10 sentences are in the final materials (3 in the development set, 7 in the formal set), involving 7 pairs in the formal set.
- **Pre-specified descriptive analyses**: the prior distribution, endpoint rates for each condition, agreement and reversal rates, between-call variability, and stratification by the original δ-SNLI split (dev/test).

## 3.6 Pre-specification, freezing, and deviations

The analysis plan (hypotheses, sidedness, SESOIs, missing-data rules, B, Holm families) was written before the formal set was run, revised over several rounds of independent review (conducted by another model), and then frozen. The plan was thus pre-specified, frozen and hashed before the formal run; it was not deposited in a public registry. The experiment fingerprint binds the plan file, the sampling, the prompts, the parsing, the run program, and a digest of the materials; any change alters the fingerprint and requires the development set to be rerun. The analysis freeze additionally binds the analysis program and B, and the formal analysis was executed only after the reconciliation (every planned unit accounted for) was complete.

All revisions made before the formal run are recorded in the plan's change log, in three rounds:

- **v0.6 (after the first development-set smoke test)**: parsing rule v2; exclusion of protected-attribute items; the criterion for codex disconnections (disconnections in the middle of sampling); and removal of the originally planned Qwen3 thinking-on configuration and its hypothesis H3 (we decided not to run it). These revisions were based on diagnostics such as format-failure types and refusals; no probability values or direction results were looked at during the revision.
- **v0.7 (after the acceptance check of the development set under the new fingerprint)**: the formal set was expanded to 240 pairs for the local configurations and 120 pairs for codex. This step **drew on exploratory diagnostics and power scenarios from the development set** (development-set movements and differences), but no formal-set results were looked at.
- **v0.7.1**: codex was switched to parallel execution; the rules were unchanged, but the fingerprint changed, and the development set was rerun.

For the original three configurations, the only deviation after the formal run began is Addendum A1 (see 3.3): it handles calls that "could not connect to the server", which is a different situation from the mid-sampling disconnections addressed in v0.6; it changes only the inclusion rules for codex and does not change any hypothesis or analysis.

**Extension (v2).** The extension kept the frozen hypotheses, SESOIs, prompts, parser and Holm families unchanged; its own task specification and harness code were bound into a separate experiment fingerprint before its development set was run. Three changes and events are recorded for transparency. (i) Two partial development-set runs were discarded and the development set was rerun from the start: once because the harness updated itself during the run (1.3.5 → 1.4.0), and once after the environment check was narrowed from a hash of the whole settings file to a hash of its permission settings, because unrelated changes to the file (made by other software) kept stopping the run. No results of the discarded runs were analysed. (ii) During the formal runs, a rule outside the fingerprint was added to clear two kinds of stop automatically: service errors that occurred before any conversation started (no answer, no tokens) and transient file-access errors; every other stop still required review. These clearances change when units are run, not which attempts are included. (iii) Before the analysis freeze, the analysis program was run once in diagnostic mode (B = 1,000) on the three configurations whose formal runs had finished, to test the pipeline; its output included point estimates for those three configurations. The only change made afterwards was a fix to the accounting step, which had counted unplanned configurations as missing; no hypothesis, threshold or estimator was changed. The confirmatory analysis reported in Section 4.6 uses B = 5,000 and was run after all four configurations were complete.

**Data and code availability.** The frozen plan (with an English reader's guide), materials, prompts, run and analysis code, all result records and run metadata, and the analysis outputs are available at https://github.com/asj252/llm-judgment-studies (folder `dnli/` the extension is in `dnli/agy-extension/`). Raw model event streams are not released because they contain local paths and service conversation identifiers; their SHA-256 hashes are recorded in the result records.

Every analysis devised only after the formal results had been seen (for example, the candidate-level readings in the Results section) is labelled as exploratory.

# 4 Results

Sections 4.1–4.5 report the three original configurations (Qwen3, OLMo-2, codex); Section 4.6 reports the four extension configurations run with the same protocol; Section 4.7 adds an exploratory analysis of candidate spread for all seven; Section 4.8 summarizes the confirmatory results.

## 4.1 Sample and completeness

All planned units of the three original configurations were fully reconciled (every planned unit accounted for) (Table 2). qwen3-off and codex had no missing units; olmo2 was missing 3 units (0.12%), all under multi-estimate elicitation, and following the pre-specified rule we computed an additional conservative interval with worst-case imputation of missing units. No attempt was truncated. The formal run of codex was rerun under a new experiment fingerprint (a hash binding the plan, sampling, prompts, parser, runner and materials) after Addendum A1 (calls that fail to connect to the server do not invalidate the session); the earlier results do not enter the analysis.

Table 2: Sample and completeness

| Configuration | Pairs | Planned units | Valid | Missing | First-attempt parse rate (v2, formal set r0) | Mean first-attempt time |
|---|---|---|---|---|---|---|
| Qwen3-8B (thinking off, local) | 240 | 2600 | 2600 | 0 | 100% | 4.8 s |
| OLMo-2-7B-Instruct (local) | 240 | 2600 | 2597 | 3 | 98.9% | 2.1 s |
| codex (gpt-6.1-sol, medium) | 120 | 1400 | 1400 | 0 | 100% | 13.4 s |

(Each pair has 10 conditions; in addition, the 20 pairs of the repeat subset were each asked once more. Parsing rule v2 was revised after the development-set smoke test according to the types of format failure, without looking at the probability values; under the old parsing rule v1, the first-attempt parse rates are 77%, 58% and 100%.)

## 4.2 Directional sensitivity (H1)

Table 3: Confirmatory tests (one Holm family {H1, H2, H4a, H4b} per configuration; intervals are two-sided 95% cluster bootstrap (resampling images) intervals on the common complete set; "Bound" is the Holm-adjusted bound nearer zero: the one-sided bound for H1, H2 and H4a, and the end of the two-sided interval nearer zero for H4b)

| Configuration | Test | n | Point estimate | 95% interval | Holm α | Bound | Judgment |
|---|---|---|---|---|---|---|---|
| Qwen3 | H1 movement (points) | 240 | 18.6 | [16.5, 20.7] | 0.0125 | 16.1 | The adjusted bound exceeds the SESOI |
| | H2 stated − reported (points) | 240 | 2.0 | [0.4, 3.6] | 0.05 | 0.6 | Direction confirmed; point estimate below the SESOI |
| | H4a endpoint-rate difference (proportion) | 240 | 0.25 | [0.21, 0.29] | 0.0167 | 0.21 | The adjusted bound exceeds the SESOI |
| | H4b multi-estimate − direct (points) | 240 | −7.9 | [−9.7, −6.2] | 0.025 | −6.0 | The adjusted bound exceeds the SESOI |
| OLMo-2 | H1 | 240 | 20.0 | [17.4, 22.7] | 0.0125 | 17.1 | The adjusted bound exceeds the SESOI |
| | H2 | 240 | 4.8 | [2.4, 7.2] | 0.0167 | 2.3 | Direction confirmed; point estimate below the SESOI |
| | H4a | 237 | 0.37 | [0.33, 0.42] | 0.025 | 0.33 | The adjusted bound exceeds the SESOI (worst-case imputation of missing units) |
| | H4b | 238 | −13.7 | [−16.1, −11.2] | 0.05 | −11.0 | The adjusted bound exceeds the SESOI (worst-case imputation of missing units) |
| codex | H1 | 120 | 25.3 | [22.6, 27.8] | 0.0125 | 22.3 | The adjusted bound exceeds the SESOI |
| | H2 | 120 | 5.4 | [3.9, 7.0] | 0.0167 | 3.8 | The point estimate reaches the SESOI, but the adjusted bound does not |
| | H4a | 120 | 0.11 | [0.075, 0.14] | 0.025 | 0.075 | The point estimate reaches the SESOI, but the adjusted bound does not |
| | H4b | 120 | −2.5 | [−3.2, −1.8] | 0.05 | −1.8 | Direction confirmed; point estimate below the SESOI |

(SESOI: 5 points for H1, H2 and H4b; 10 percentage points for H4a. All 12 tests reject the null hypothesis under Holm correction, and there are no results in the opposite direction. H4b is signed as "multi-estimate < direct", and the bound is taken at the end nearer zero. For olmo2, the H4a and H4b judgments use the conservative bounds after worst-case imputation of missing units.)

Under stated presentation and direct elicitation, all three configurations moved in the direction annotated for the update, and for all three the Holm-adjusted bound exceeds the smallest effect size of interest (SESOI) of 5 points (Table 3; Figure 1): the directional movement $s$ is 18.6, 20.0 and 25.3 points. On average, strengtheners raised the probability by 15.9, 14.4 and 25.9 points, and weakeners lowered it by 21.3, 25.6 and 24.7 points; the difference D between the posteriors after strengtheners and after weakeners is 37.2, 40.1 and 50.6 points, respectively.

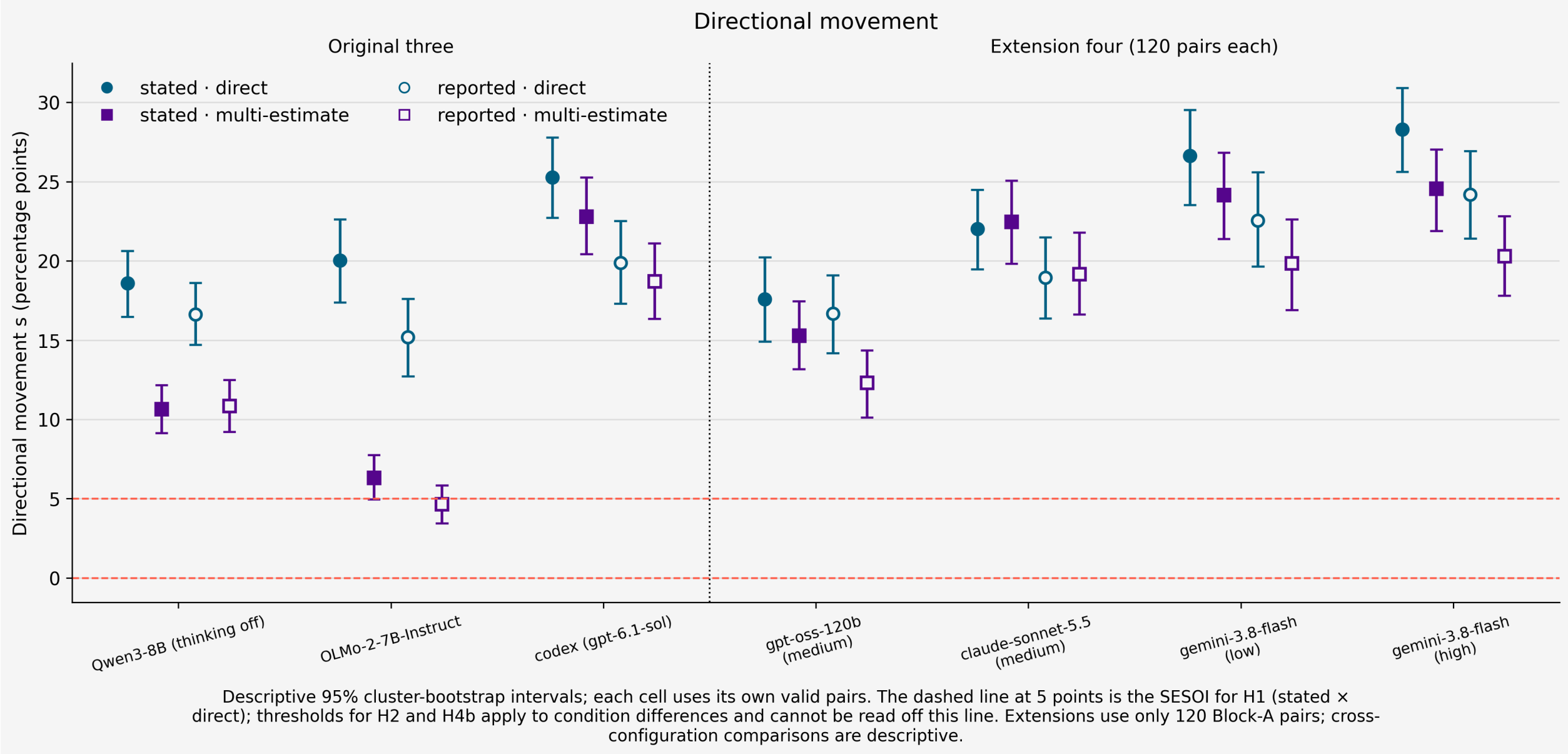

Figure 1: Directional movement s by configuration, presentation and elicitation (descriptive 95% cluster-bootstrap intervals). Original three configurations and four extension configurations are separated; each extension uses only 120 Block-A pairs.

Item by item, the proportion of moves whose direction agrees with the annotation is 0.66, 0.55 and 0.83, and the proportion of reverse moves is 0.15, 0.16 and 0.09; the remainder report the same number before and after (qwen3-off: 23% for strengtheners, 16% for weakeners; olmo2: 29% for both; codex: 9% and 8%). That is, counted by update condition, 16%–29% of update sentences left the reported probability of the local models unchanged.

## 4.3 Reporting wording (H2)

When the same update was rephrased as "Someone said that …", the directional movement became smaller in all three configurations: stated minus reported is 2.0, 4.8 and 5.4 points (Table 3). All three pass the directional test under Holm correction. For codex the point estimate reaches the 5-point SESOI, but the Holm-adjusted bound (3.85) does not exceed it; for the two local configurations the point estimate is below the SESOI.

The **reversal rate** under reported presentation is higher than under stated presentation in all three configurations: qwen3-off 0.146 → 0.173, olmo2 0.163 → 0.221, codex 0.092 → 0.117. Following the pre-specified interpretation rule, we state the conclusion as **reporting shifts the direction of movement**: after the update is recast as a report, movement in the annotated direction is smaller, and at the same time more items move in the opposite direction. This is not the same as "discounting within the same direction."

Broken down by update type (descriptive; no per-type tests were run, and there is no evidence of equivalence), the three configurations show different patterns (Table 4): for olmo2 the difference lies almost entirely in strengtheners (stated +14.4, reported +4.7), while the means for weakeners are close (−25.6 vs. −25.7; equivalence not tested); for codex the difference lies mainly in weakeners (−24.7, −16.5); for qwen3-off both types decrease only slightly. This is a post hoc observation, offered only as exploratory description.

Table 4: Mean probability change by update type (direct elicitation; points)

| Configuration | Strengthener, stated | Strengthener, reported | Weakener, stated | Weakener, reported |
|---|---|---|---|---|
| Qwen3 (thinking off) | +15.9 | +15.0 | −21.3 | −18.3 |
| OLMo-2 | +14.4 | +4.7 | −25.6 | −25.7 |
| codex | +25.9 | +23.3 | −24.7 | −16.5 |

## 4.4 Multi-estimate elicitation (H4a, H4b)

**Endpoints (H4a).** Under direct elicitation, a large share of responses fall at ≤5% or ≥95% (Figure 2 shows the unweighted rate in each condition): the endpoint rate, computed with the pre-specified weights (prior 1/2, stated strengthener 1/4, stated weakener 1/4), is 29.6%, 42.0% and 25.2%. With multi-estimate elicitation (listing 5 estimates, each with a weight, and taking the weighted mean), it drops to 4.3%, 4.9% and 14.4%. The differences 0.25, 0.37 and 0.11 all pass Holm correction; for the two local configurations the adjusted bound exceeds the SESOI of 10 percentage points, while for codex the point estimate reaches the SESOI but the adjusted bound does not (Table 3).

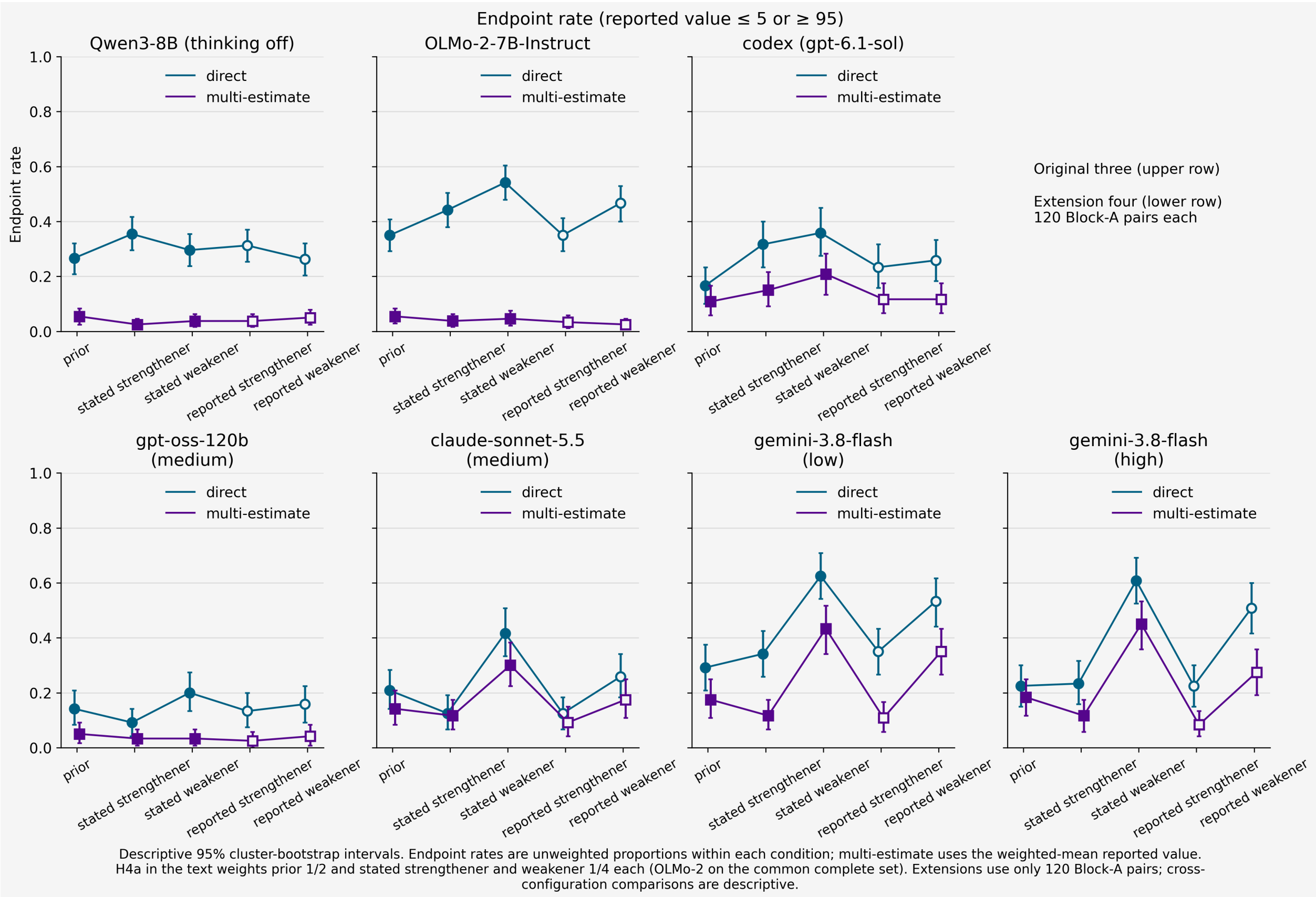

Figure 2: Endpoint rate (reported value ≤ 5 or ≥ 95) by condition and elicitation. The original three configurations are in the upper row; the four extensions are in the lower row and each uses only 120 Block-A pairs.

**Directional sensitivity (H4b).** Multi-estimate elicitation reduces the directional movement: multi-estimate minus direct is −7.9, −13.7 and −2.5 points. For the two local configurations the adjusted bound lies beyond −5 points (the whole interval falls below −5 points); for codex the direction is confirmed and the point estimate is below the SESOI (Table 3).

Taken together, multi-estimate elicitation reduced endpoint values in all three configurations; in the two local models it also reduced the mean directional movement, by about 43% (qwen3-off 18.6 → 10.7) and 68% (olmo2 20.0 → 6.3, 238 pairs), respectively; for codex the directional movement decreased by about 2.5 points (25.3 → 22.8), with the direction confirmed and the point estimate below the 5-point SESOI.

Viewed by update condition, the change in the local models is not merely a smaller magnitude: under multi-estimate elicitation, **the proportion of zero movement falls and the proportion of reverse movement rises**. Zero movement: qwen3-off strengtheners 22.9% → 4.6%, weakeners 16.3% → 5.8%; olmo2 29.2% → 0.4%, 29.2% → 1.3%. Reverse movement: qwen3-off 0.14 → 0.34, 0.15 → 0.24; olmo2 0.17 → 0.41, 0.16 → 0.40. The pooled agreement rate, however, did not fall (qwen3-off 0.658 → 0.658, olmo2 0.546 → 0.586): update sentences that originally produced no movement moved, under multi-estimate elicitation, some in the annotated direction and some in the opposite direction. For codex the reversal rate is similar to that under direct elicitation (0.067 → 0.067, 0.117 → 0.125), and the pooled agreement rate goes from 0.825 to 0.871.

**Candidate level (descriptive).** Under the stated condition, the weighted standard deviation of the 5 estimates is 16.1 points for qwen3-off, 18.3 points for olmo2 and 7.6 points for codex. The proportion of responses whose weights sum to a value in [0.95, 1.05] is 93–94% for qwen3-off, 22–29% for olmo2 and 100% for codex (non-conforming weights are normalized by their sum; the weighted mean is invariant to overall scaling).

**Readings within the candidates (exploratory, not pre-specified).** To distinguish "the weighted mean pulls dispersed candidates toward the middle" from "the candidates themselves move little", we recomputed the directional movement post hoc with two other readings: the top-weighted candidate and the weighted median (Table 5). For qwen3-off, switching to the top-weighted candidate shrinks the difference from direct elicitation from −7.9 to −4.6 points (a reduction of about 42%), and the directional movement of the top-weighted candidate remains about 14.0 points; for olmo2 the top-weighted-candidate difference is −14.5 points, larger than that of the weighted mean, and its directional movement is only about 5.5 points; for codex all three readings are around −2 points. This describes only the size of the differences under different readings and does not support attributing a mechanism: model capability, vendor, training method and CLI harness are confounded with one another, and the present data also cannot separate the two explanations "candidates interfere with one another" and "format burden". These analyses were conceived only after we had seen the formal results and are offered only as exploration.

Table 5: Directional movement under the three multi-estimate readings minus direct elicitation (stated; points; brackets give 95% cluster bootstrap (resampling images) intervals, 2000 resamples, seed 20261490, from the same computation as Figure 3)

| Configuration | Candidate weighted SD (stated) | Weighted mean | Top-weighted candidate | Weighted median |
|---|---|---|---|---|
| Qwen3 (thinking off) | 16.1 | −7.9 [−9.7, −6.2] | −4.6 [−6.4, −2.7] | −6.7 [−8.5, −4.9] |
| OLMo-2 | 18.3 | −13.7 [−16.1, −11.3] | −14.5 [−17.0, −11.9] | −12.8 [−15.3, −10.2] |
| codex | 7.6 | −2.5 [−3.2, −1.8] | −2.1 [−2.8, −1.4] | −2.3 [−3.0, −1.7] |

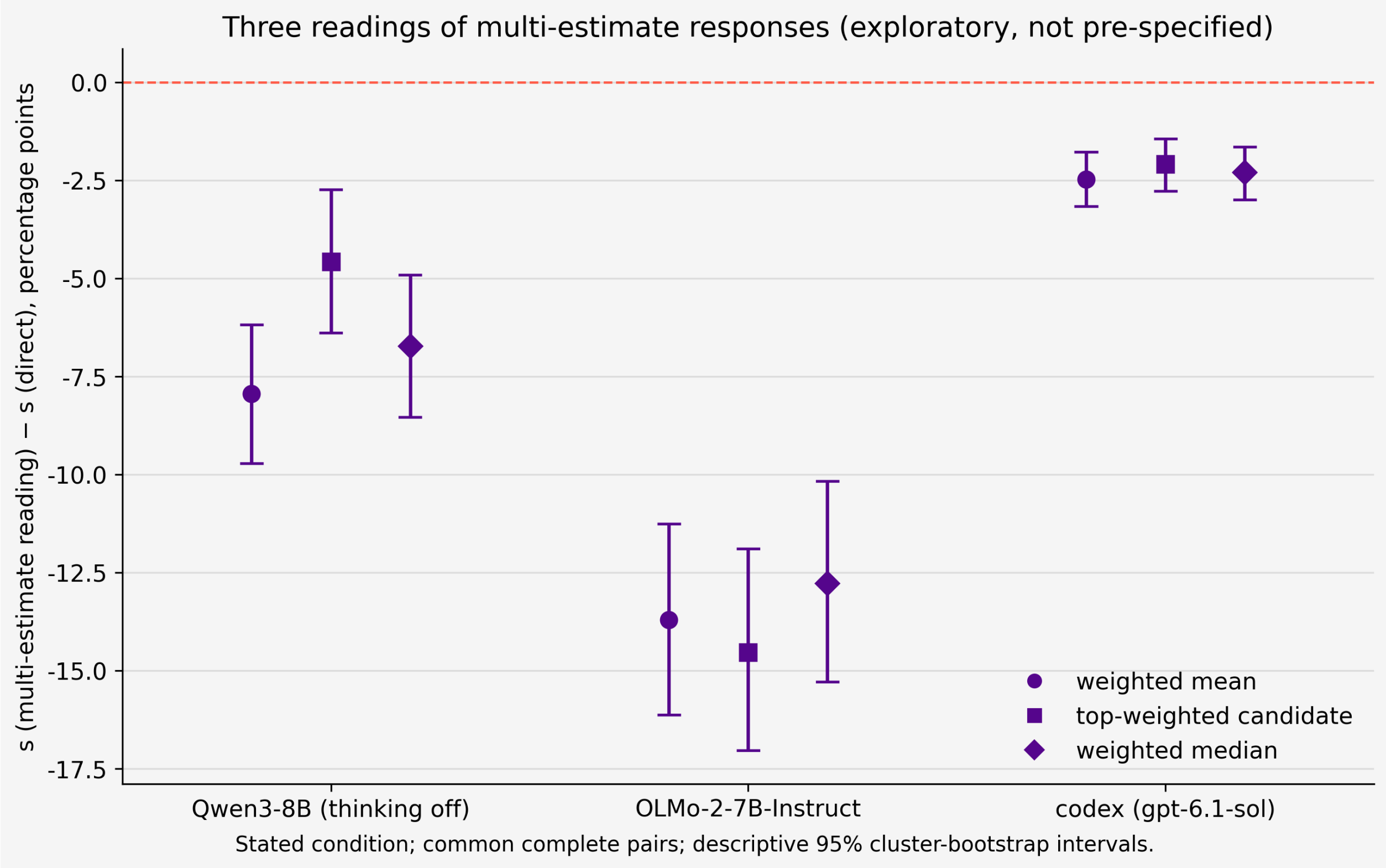

Figure 3: Multi-estimate readings minus direct elicitation, stated condition (exploratory, not pre-specified).

## 4.5 Robustness and supplementary description

- **Quoted-speech updates** (updates whose reported version becomes quoted speech embedded in a report; pre-specified; of the 11 IDs on the frozen list, the formal set contains 7 sentences, involving 7 pairs): after excluding these 7 pairs and recomputing, the H1, H2 and H4b conclusions are unchanged in all three configurations (e.g., H2: 2.05, 4.92, 5.48 points; codex was run only on Block A, leaving 117 pairs after exclusion).
- **The common 120 pairs**: on the same 120 pairs used for codex, the two local configurations' point estimates for H1, H2, H4a and H4b have the same direction as on the 240 pairs, although the values differ (e.g., olmo2 H1 22.0 vs. 20.0, H2 6.2 vs. 4.8; for qwen3-off, H2 is 1.7 points, with interval [−0.6, 3.9] spanning 0). Cross-configuration comparisons are descriptive only.
- **Stratification by material source** (the dev and test splits of δ-SNLI, unrelated to this study's development set): the direction is consistent across the two strata; H2 is larger on items from dev in all three configurations.
- **Between-call variability** (repeat subset of 20 pairs): when the same prompt was asked a second time, the mean absolute difference between the two reported values was 9.6, 7.7 and 2.3 points under direct elicitation, and 7.7, 9.5 and 2.0 points under multi-estimate elicitation. Under stated presentation, the mean absolute difference in $s$ is 5.7, 7.2 and 2.3 points, whereas the between-item standard deviation is 16.9, 23.7 and 15.3 points. Single reported values from the local models are very noisy; the conclusions rest on averages over 240 pairs and are not suited to interpreting individual items.
- **Comparison with the development set**: on the development set (30 pairs), H2 was 4.3, 4.6 and 5.7 points (7.0 for codex on the development set rerun after Addendum A1); on the formal set, qwen3-off is about half its development-set value, and the others are close. The H4b pattern of "large differences for the local models, a small difference for codex" recurs in the formal set.

## 4.6 Extension: four further configurations (v2)

The four extension configurations (Section 3.3) were run on the same 120 Block-A pairs and the same 20-pair repeat subset as codex, with the same prompts, parser (with one input normalization; Section 3.3), hypotheses, SESOIs and Holm families; each configuration forms its own family {H1,

H2, H4a, H4b}. All planned units were reconciled and none were missing (Table 6). Retries were rare: in the final records, 5, 5, 8 and 5 earlier attempts were discarded, almost all because the event stream contained a step other than the answer (classified as possible tool use and resampled), and three because of a timeout; counted by the run's persistent retry counter, which also covers sessions that were later resumed, gemini-3.8-flash (low) had 10 retries and the others 5. According to the run log, six units of gemini-3.8-flash (low) were blocked at least once by the provider's content filter, which surfaced as such a step; one of them was blocked three times in a row and stopped the run. On review (recorded in the run's review log), the stop was cleared once so that the unit could be retried, because blocks appeared to be stochastic (other conditions of the same pair had passed on retry); we decided at the same time that units blocked persistently would be recorded as missing and reported in an addendum. The retried unit was answered on its next attempts, so this provision was never needed. All six units therefore have valid answers, and no unit is missing for this reason. Service errors that occurred before a conversation started also stopped the run and were cleared under a written rule before resuming (Section 3.6); such calls produced no answer and leave no attempt in the analysed record.

Table 6: Sample and completeness (extension)

| Configuration | Pairs | Planned units | Valid | Missing | First-attempt success (formal set r0) | Mean first-attempt time |
|---|---|---|---|---|---|---|
| gpt-oss-120b (medium) | 120 | 1400 | 1400 | 0 | 99.7% | 9.1 s |
| claude-sonnet-5.5 (medium) | 120 | 1400 | 1400 | 0 | 99.7% | 10.2 s |
| gemini-3.8-flash (low) | 120 | 1400 | 1400 | 0 | 99.4% | 9.6 s |
| gemini-3.8-flash (high) | 120 | 1400 | 1400 | 0 | 99.6% | 19.1 s |

Table 7: Confirmatory tests (extension; same layout as Table 3; n = 120 for every test; "—" = not rejected under Holm, so no bound is reported)

| Configuration | Test | Point estimate | 95% interval | Holm α | Bound | Judgment |
|---|---|---|---|---|---|---|
| gpt-oss-120b | H1 movement (points) | 17.6 | [15.0, 20.2] | 0.0125 | 14.5 | The adjusted bound exceeds the SESOI |
| | H2 stated − reported (points) | 0.9 | [−0.9, 2.7] | 0.05 | — | Not rejected; direction not established |
| | H4a endpoint-rate difference (proportion) | 0.10 | [0.067, 0.142] | 0.0167 | 0.065 | The point estimate reaches the SESOI, but the adjusted bound does not |
| | H4b multi-estimate − direct (points) | −2.3 | [−4.1, −0.6] | 0.025 | −0.3 | Direction confirmed; point estimate below the SESOI |
| claude-sonnet-5.5 | H1 | 22.0 | [19.4, 24.6] | 0.0125 | 19.0 | The adjusted bound exceeds the SESOI |
| | H2 | 3.1 | [1.8, 4.5] | 0.0167 | 1.7 | Direction confirmed; point estimate below the SESOI |
| | H4a | 0.065 | [0.038, 0.094] | 0.025 | 0.038 | Direction confirmed; point estimate below the SESOI |
| | H4b | +0.5 | [−0.3, 1.3] | 0.05 | — | Not rejected; direction not established |
| gemini-3.8-flash (low) | H1 | 26.6 | [23.6, 29.6] | 0.0125 | 23.2 | The adjusted bound exceeds the SESOI |
| | H2 | 4.1 | [2.4, 5.9] | 0.0167 | 2.2 | Direction confirmed; point estimate below the SESOI |
| | H4a | 0.16 | [0.123, 0.204] | 0.025 | 0.12 | The adjusted bound exceeds the SESOI |
| | H4b | −2.5 | [−3.7, −1.4] | 0.05 | −1.4 | Direction confirmed; point estimate below the SESOI |
| gemini-3.8-flash (high) | H1 | 28.3 | [25.6, 31.0] | 0.0125 | 25.1 | The adjusted bound exceeds the SESOI |
| | H2 | 4.1 | [2.7, 5.8] | 0.0167 | 2.5 | Direction confirmed; point estimate below the SESOI |
| | H4a | 0.090 | [0.056, 0.123] | 0.025 | 0.056 | Direction confirmed; point estimate below the SESOI |
| | H4b | −3.7 | [−5.1, −2.5] | 0.05 | −2.5 | Direction confirmed; point estimate below the SESOI |

(No unit was missing, so no worst-case imputation was needed. No result fell in the direction opposite to a one-sided prediction.)

**Directional sensitivity (H1).** All four configurations moved in the annotated direction, and for all four the adjusted bound exceeds the 5-point SESOI: $s = 17.6$, 22.0, 26.6 and 28.3 points. The difference between the posteriors after a strengthener and after a weakener, D, is 35.2, 44.0, 53.3 and 56.6 points. Zero movement under stated presentation was less common than in the local models (6.7%–15.0% of update conditions, against 16%–29%).

**Reporting wording (H2).** Recasting the update as a report reduced the directional movement in three configurations (3.1, 4.1 and 4.1 points; direction confirmed, point estimates below the SESOI); for gpt-oss-120b the difference was 0.9 points and the direction was not established. As in the original three configurations, the reversal rate was higher under reported than under stated presentation in all four (0.163 → 0.192, 0.079 → 0.125, 0.062 → 0.092 and 0.050 → 0.079), so under the pre-specified rule the conclusion is again worded as "reporting shifts the direction of movement". By update type (descriptive), the reduction lay mainly in weakeners for claude-sonnet-5.5 (−18.6 → −13.9) and gemini-3.8-flash (low) (−26.4 → −19.6), and was spread over both types for gemini-3.8-flash (high) (+27.9 → +24.9; −28.7 → −23.5).

**Multi-estimate elicitation (H4a, H4b).** The weighted endpoint rate under direct elicitation was 14.4%, 24.0%, 38.8% and 32.3%, and under multi-estimate elicitation 4.2%, 17.5%, 22.5% and 23.3%. The reduction passed Holm correction in all four configurations; only for gemini-3.8-flash (low) does the adjusted bound exceed the 10-point SESOI. The change in directional sensitivity was small: −2.3, +0.5, −2.5 and −3.7 points (direction confirmed in three configurations, point estimates below the SESOI; not established for claude-sonnet-5.5). The pattern seen in the local models recurs on a smaller scale: under multi-estimate elicitation, zero movement almost disappears (at most 3.3% of update conditions), the reversal rate rises (pooled: 0.163 → 0.212, 0.079 → 0.117, 0.062 → 0.133, 0.050 → 0.113), and the pooled agreement rate does not fall (0.746 → 0.783, 0.846 → 0.875, 0.804 → 0.846, 0.846 → 0.887). The weighted standard deviation of the five candidates under stated presentation was 14.0 points for gpt-oss-120b and 6.3–7.5 points for the other three; every multi-estimate response had weights summing to a value in [0.95, 1.05].

**Robustness (pre-specified).** Excluding the three Block-A pairs with quoted-speech updates (117 pairs remaining) leaves every conclusion unchanged (e.g., H2: 0.85, 3.07, 4.16 and 4.19 points). Stratified by material source, H2 was larger on δ-SNLI dev items for two configurations and on test items for the other two, unlike the original three configurations, where it was larger on dev items in all three. Between-call variability was small except for gpt-oss-120b: on the repeat subset, the mean absolute difference in $s$ between two calls of the same prompt was 5.9, 1.4, 2.3 and 1.4 points, against between-item standard deviations of 17.5, 14.5, 15.5 and 14.7 points.

**Reasoning effort (descriptive only).** The two gemini-3.8-flash configurations differ only in reasoning effort. On the common 120 pairs, high minus low effort was +1.7 points for H1 (descriptive 95% interval [0.3, 3.2]), 0.0 for H2 ([−1.9, 2.0]), −0.073 for H4a ([−0.110, −0.037]) and −1.3 for H4b ([−2.7, 0.1]). These contrasts are not part of any Holm family.

## 4.7 Candidate spread (exploratory, not pre-specified)

Under multi-estimate elicitation each response gives five candidate probabilities with weights, so it also carries a spread: the weighted standard deviation of the candidates (weights normalized to sum to 1; $SD = \sqrt{\sum w_i (x_i - m)^2}$, where m is the weighted mean). We asked three questions after seeing the formal results: does the spread shrink after an update, is it larger after a report than after a stated update, and do items with a more dispersed prior move more? Intervals are cluster-bootstrap 95% intervals (B = 2,000, seed 20261500); none of these analyses enters a Holm family.

Table 8: Candidate spread (stated and reported posteriors versus prior; points, with the change as a percentage of the reference spread)

| Configuration | Prior spread | Stated posterior minus prior | Reported minus stated | Spearman ρ (prior spread, $s$) |
|---|---|---|---|---|
| Qwen3 (thinking off) | 15.8 | +0.3 (+2%) | −2.1 (−13%) [−2.8, −1.5] | 0.09 [−0.04, 0.21] |
| OLMo-2 | 17.5 | +0.8 (+4%) | 0.0 (0%) [−0.8, 0.9] | 0.10 [−0.03, 0.22] |
| codex | 9.2 | −1.6 (−18%) | +0.7 (+9%) [0.3, 1.0] | 0.15 [−0.05, 0.33] |
| gpt-oss-120b | 15.7 | −1.7 (−11%) | +0.9 (+7%) [0.1, 1.7] | 0.08 [−0.11, 0.25] |
| claude-sonnet-5.5 | 7.0 | −0.7 (−10%) | +0.6 (+9%) [0.3, 0.9] | 0.31 [0.13, 0.48] |
| gemini-3.8-flash (low) | 9.0 | −1.5 (−17%) | +0.4 (+6%) [−0.01, 0.86] | 0.24 [0.04, 0.42] |
| gemini-3.8-flash (high) | 8.8 | −1.3 (−15%) | +0.8 (+11%) [0.4, 1.3] | 0.26 [0.08, 0.43] |

(“Stated posterior minus prior” averages the strengthener and weakener changes; its percentage is relative to the prior spread. “Reported minus stated” pools strengtheners and weakeners; its percentage is relative to the stated-posterior spread.)

The two local models show no consistent pattern: in Qwen3 a stated strengthener widened the spread (+2.6 points) and a weakener narrowed it (−2.1), so the average barely moves, and after a report the spread was smaller, not larger; in OLMo-2 the changes were near zero. The five configurations accessed through agent harnesses behave alike: after a stated update the spread narrowed by 10%–18% of the prior spread (mainly after weakeners in the four extension configurations), and after a report it was 6%–11% wider than after the same update stated (the interval excludes zero in four of the five). In claude-sonnet-5.5 and both gemini-3.8-flash configurations, items with a more dispersed prior also moved more (ρ = 0.24 to 0.31); in the other four configurations the correlation was weak (0.08 to 0.15) and its interval included zero. These are small absolute differences (under 2 points of spread) and were examined only post hoc. In the harnessed configurations the candidate spread thus shows the pattern one would expect of an expression of uncertainty (narrowing with information, staying wider for hearsay), whereas in the two small local models it does not show this pattern; there is no independent measure of uncertainty against which to check either, and the present data cannot say whether this reflects model scale, the harnesses, or training.

### 4.8 Summary

1. All seven configurations move in the annotated direction in response to a single-sentence update; the directional movement ranges from about 18 to 28 points, and for every configuration the adjusted bound exceeds the 5-point SESOI.
2. Recasting the update as an anonymous report reduces the movement in all seven point estimates (0.9 to 5.4 points) and raises the reversal rate in all seven. The direction is confirmed in six configurations (2.0 to 5.4 points); for gpt-oss-120b it is not established. The point estimate reaches 5 points only for codex, and no adjusted bound exceeds 5 points.
3. Multi-estimate elicitation reduces endpoint values in all seven configurations (the adjusted bound exceeds 10 percentage points in three). Its cost in directional sensitivity is large in the two local models (−7.9 and −13.7 points, adjusted bounds beyond −5); in the five configurations accessed through agent harnesses the point estimates lie between −3.7 and +0.5 points, the direction is confirmed in four, and no clear change is detected for claude-sonnet-5.5. Wherever zero movement was common under direct elicitation, multi-estimate elicitation replaced it with a mixture of agreeing and reversed movement.

## 5 Discussion

### 5.1 Main findings

Under single-sentence, direction-controlled updates, the probabilities reported by all seven model configurations moved in the annotated direction, with a directional movement of roughly 18 to 28 points. When the same update was recast as an anonymous report, the point estimate of the movement became smaller in all seven configurations (significantly in six), and in all seven more items moved in the opposite direction; no point estimate exceeded 5.4 points. Switching to the elicitation that asks for “5 estimates, each with a weight” reduced endpoint values in all seven configurations; on the two local models it also clearly reduced the response to the update, whereas for the five configurations accessed through agent harnesses the point estimates of the reduction were below 4 points (no clear change for claude-sonnet-5.5).

### 5.2 Directional sensitivity: the models move, but not every time

H1 shows that these models do not ignore new information: strengtheners raised the probability and weakeners lowered it, and the posterior difference D between strengtheners and weakeners was 35 to 57 points across the seven configurations. Item by item, however, 16% to 29% of update sentences left the reported probability unchanged for the local models, and 7% to 15% did so for codex and the four extension configurations. One possibility is that, under direct elicitation, the models tend to report a few common round numbers; when the prior and the posterior fall on the same round number, no movement can be measured. The between-call variability reported in the Results section (for the local models, two reports for the same prompt differed by 8 to 10 points on average) is also a reminder that a single reported value is coarse, and that conclusions must rest on averages over many pairs.

We do not claim that these movements are “too large” or “too small.” δ-NLI annotates only direction and provides no reference for an appropriate magnitude; answering the magnitude question would require people to report probabilities under the same paradigm (see 5.7).

### 5.3 Reporting wording: smaller movement, more reversals

Replacing “Here is something else you learned” with “Someone said that” reduced the directional movement in all three original configurations (by 2.0 to 5.4 points) and in three of the four extension configurations (by 3.1 to 4.1 points); for gpt-oss-120b the difference was 0.9 points and its direction was not established. If the report merely led the model to discount the source somewhat, one simple expectation would be a proportional shrinkage of movement in the original direction; but we also observed a rise in the reversal rate (e.g., from 0.16 to 0.22 for olmo2). The reversal rate was higher under the report in all seven configurations, including gpt-oss-120b. Following the pre-specified rule, we therefore state this result as "reporting shifts the direction of movement." Discounting as a mechanism was neither tested nor ruled out in this study. Only for codex does the point estimate reach the 5-point smallest effect size of interest (SESOI), and for no configuration does the adjusted bound exceed it. Across seven configurations, then, the models register that a statement is a report, but the resulting change in how far they move is small relative to the movement itself: the point estimates correspond to a reduction of roughly 5% to 25% of the stated movement.

One explanation is pragmatic: in everyday conversation, going out of one's way to say "someone said p" sometimes implies that the speaker does not fully believe p; if the models have learned this pragmatic implication, they may treat the report as expressing some reservation about p, or even as a cue against it. Another explanation is that "Someone said that" turns the update into the event of "someone having said something," so that the model instead reasons about the speech situation rather than about the content of the update. Linguistic work on the extent to which "X said that p" licenses p (Ross & Pavlick, 2019; Jiang & de Marneffe, 2019) is compatible with both explanations, but the dependent variable in that work is an entailment judgment, not the probability movement for the same hypothesis.

Broken down by update type, the configurations show different patterns: for olmo2 the difference lies almost entirely in strengtheners, whereas for codex, claude-sonnet-5.5 and gemini-3.8-flash (low) it lies mainly in weakeners. This is a post hoc description, and no per-type test was conducted; it suggests that reporting may act in different ways on different models, which merits testing with more reporting wordings (e.g., named sources, sources of differing credibility).

## 5.4 Multi-estimate elicitation: de-extremizing and desensitizing

Multi-estimate elicitation achieved one of its expected effects: endpoint values fell sharply (for the local models, the weighted endpoint rate dropped from about 30%–42% to about 5%). This matters for research on probability updating: if the prior is already reported as 100%, a strengthener has no room to move it upward.

On the local models, however, the same elicitation also clearly reduced the directional movement (by about 43% and 68%). Item by item, the change has two sides: most update sentences that produced "no movement at all" under direct elicitation began to move, but a substantial share of them moved in the reverse direction, so the pooled agreement rate neither declined nor rose by much. The picture for the configurations accessed through agent harnesses is different: for codex the endpoint rate fell from 25% to 14% and the directional movement decreased by only about 2.5 points; for the four extension configurations the directional movement changed by between −3.7 and +0.5 points. In these five configurations the item-level pattern of the local models recurs on a smaller scale (zero movement nearly disappears, reversals rise, and the pooled agreement rate does not fall), but the point estimates of the net cost in directional sensitivity are all below the 5-point SESOI (the intervals do not exclude larger costs in every case; for gemini-3.8-flash (high) the 95% interval reaches −5.1). For the two gemini-3.8-flash configurations, which differ only in reasoning effort, the descriptive differences (high minus low) were +1.7 points for H1, 0.0 for H2, −0.073 for H4a and −1.3 points for H4b; no equivalence test was made.

This raises a question our data cannot yet answer: **does multi-estimate elicitation make the models' judgments more "correct," or does it merely make them respond somewhat less to everything?** In δ-NLI, the update **should** push the estimate in a given direction, but there is no reference for an appropriate magnitude, so a smaller average directional movement cannot yet be judged as an improvement or a decline in judgment quality. In other settings, information **should not** move the estimate (e.g., an irrelevant numerical anchor, or a known outcome when estimating the ex ante probability); if multi-estimate elicitation also reduced responses there, it would look like debiasing. Only by placing the two kinds of settings side by side can one tell whether multi-estimate elicitation leads models to "move when they should and stay put when they should not," or whether it simply dampens responses across the board. We list this test as future work (5.7).

The exploratory analysis in the Results section also shows that the problem does not lie only in "averaging pulls dispersed candidates toward the middle": olmo2′s top-weighted candidate also moved by only about 5.5 points on average. But readings of this kind were conceived only post hoc, and the configurations are mutually confounded in model capability, vendor, training method, and calling method (local inference versus agent harnesses), so we cannot use them to say anything about "weaker models." The contrast between the two local configurations and the five harnessed configurations is also a contrast between small open models run locally and larger models run through vendors' harnesses; the present design cannot separate these. Nor can the present data separate two explanations: that listing multiple candidates itself interferes with the model's use of the evidence, or that the format consumes the model's capacity (e.g., among olmo2′s parsable responses, only 22%–29% had weights summing to within [0.95, 1.05]).

The spread of the five candidates tells a related story (Section 4.7, exploratory). In the five harnessed configurations the spread narrowed after an update and stayed somewhat wider after a report than after the same statement, which is what one would expect if the candidates expressed uncertainty; in the two local models it did not. So the five estimates are not a uniform readout of uncertainty across models, and studies that use verbalized candidate spread as an uncertainty measure should check this per model.

The original Verbalized Sampling paper discusses gains in diversity and reports that stronger models benefit more (Zhang et al., 2025); the present study measures directional sensitivity, and the two are different dependent variables. Whether the effect of multi-estimate elicitation on directional sensitivity varies with model capability needs to be tested directly, after controlling for vendor and calling method.

## 5.5 Implications for using multi-candidate elicitation

Multi-candidate formats such as Verbalized Sampling are attractive because they reduce extreme answers and expose a distribution rather than a single number. Our results add a caution and a tool. The caution: the same format had very different side effects across models. In the two small local models it cut the response to evidence by roughly half and its candidate spread did not show the pattern expected of an uncertainty expression; in the five harnessed configurations the point estimates of the cost were small (though the interval for gemini-3.8-flash (high) reaches −5.1 points) and the spread showed that pattern. A lower endpoint rate, on its own, therefore does not show that a format has made a model's judgments better. The tool: the present protocol is a cheap behavioral diagnostic that can be run before adopting such a format for a given model. It requires only items whose evidence has a known direction, two separate calls per item, and the two elicitation formats; its outputs (endpoint rate, directional movement, reversal and zero-movement rates, and candidate spread before and after evidence) form a response profile that can be compared across models. We use "profile" descriptively: the present design cannot attribute differences between profiles to model scale, vendor, training or harness, and it covers a single task. Whether the profiles generalize to other judgment tasks is the subject of the follow-up study mentioned in 5.7.

## 5.6 Limitations

- **Materials**: δ-SNLI was released in 2020 and may already be present in training corpora; the effect of contamination on each condition is unknown. The direction of an update reflects the writer's intent; in the original paper, about 8.9% of weakeners were judged by validators to be strengtheners. Our direction spot check involved only one human rater (the author), covered 20 items, and examined only the stated version. The premises come from image captions and tend to be short.

- **Wording**: Reporting used only one wording ("Someone said that"), and the lead-in sentence of the stated version itself signals that the information is true; H2 compares the total effect of these two wordings, and its conclusions are limited to these two wordings.
- **Measurement**: The prior and the posterior come from two independent calls; $\Delta$ is "the difference between the probabilities reported under two inputs," not a revision within a single reasoning process. Single reported values from the local models are noisy. Parsing rule v2 was revised after seeing the format failures on the development set (the probability values were not examined during the revision).
- **Models**: There are seven configurations of six named models; capability, vendor, training method, and calling method are mutually confounded. Five configurations were called through agent harnesses (the codex CLI and Google's agy CLI) whose system prompts, sampling parameters and server-side versions are outside our control, so conclusions apply only to "the model within the harness"; the extension was run about a week after the original configurations, and server-side updates within that period cannot be ruled out (the harness version, settings and reported model name were monitored). The two local models used the same set of sampling parameters, and the effect of settings such as temperature was not tested. The extension used only the 120 Block-A pairs, which limits the precision of its estimates.
- **Reference**: Without human probabilities under the same paradigm, we cannot judge whether movement is excessive or insufficient.

### 5.7 Future work

1. **Human reference**: Ask people to report priors and posteriors on the same set of pairs (multiple raters per item), to answer the magnitude question and to test whether the reporting wording also "shifts the direction of movement" for humans.
2. **Reasoning effort and capability**: The extension compared two reasoning efforts of one model only descriptively. A design that varies effort within several models, and capability within one vendor, is needed to say whether the cost of multi-estimate elicitation depends on either.
3. **"Should move" versus "should not move"**: Carry multi-estimate elicitation over to classic paradigms in which information should not move the estimate, such as anchoring (an irrelevant number) and hindsight bias (estimating the ex ante probability when the outcome is known), as well as base-rate problems, in which information should be used but is often neglected. Our preliminary literature search did not find studies that apply the "verbalize multiple candidates at once" elicitation to these paradigms. A follow-up study along these lines is in preparation.
4. **Variants of reporting**: Named sources, sources of differing credibility, and direct versus indirect quotation, to separate the two explanations of pragmatic implication and source credibility.